\documentclass[11pt]{article}
\usepackage{acl2015}
\usepackage{times}
\usepackage{url}
\usepackage{latexsym}
\usepackage{todonotes}
\usepackage{natbib}
\usepackage{appendix}
\usepackage{afterpage}

\graphicspath{ {graphics/} }

\title{Topic Modeling in the Voynich Manuscript\thanks{Thanks to Luke Lindemann, Lisa F. Davis, and members of the Yale class ``Linguistics of the Voynich Manuscript'' in 2019. This paper began as a project with 5 members of the class, including RS and AP. This article was written by RS and CB, using analyses mostly by RS, with contributions by AP.}}

\begin{small}
\author{Rachel Sterneck \\
  Yale University \\
  Department of Computer Science \\
  New Haven, CT 06520 \\
  {\tt rachel.sterneck@yale.edu} \\\And
 
 Annie Polish \\
  Yale University \\
  { } \\
  New Haven, CT 06520 \\
  {\tt annie.polish@yale.edu} \\\And
 
  Claire Bowern \\
  Yale University  \\
  Department of Linguistics \\
   New Haven, CT 06520 \\
  {\tt claire.bowern@yale.edu} \\}
\end{small}
\date{\today}

\begin{document}
\maketitle
\begin{abstract}\begin{small}
This article presents the results of investigations using topic modeling of the Voynich Manuscript (Beinecke MS408). Topic modeling is a set of computational methods which are used to identify clusters of subjects within text. We use latent dirichlet allocation, latent semantic analysis, and nonnegative matrix factorization to cluster Voynich pages into `topics'. We then compare the topics derived from the computational models to clusters derived from the Voynich illustrations and from paleographic analysis. We find that computationally derived clusters match closely to a conjunction of scribe and subject matter (as per the illustrations), providing further evidence that the Voynich Manuscript contains meaningful text.\end{small}
\end{abstract}

\section{Introduction}
The Voynich Manuscript (Beinecke Library MS 408)\footnote{High resolution images of all pages of the manuscript are available from \texttt{https://brbl-dl.library.yale.edu/ vufind/Record/3519597}. The text is available in machine-readable format from \texttt{voynich.nu}. Our scripts and data are available from \texttt{https://github.com/rachelsterneck/voynich-topic-modeling}.}  has puzzled linguists, historians, and conspiracy theorists alike for its unrecognizable text and varied illustrations. For a medieval manuscript, it is quite surprising that no one has been able to identify its language of origin or break its cipher; this has prompted many to believe that the manuscript is a hoax \citep[e.g.][]{rugg2004,barlow1986,TimmSchinner2020}.

However, statistical approaches offer a novel way of analyzing the Voynich blackbox. Statistical methods offer tools that capture relevant features of the text without understanding its meaning, and more importantly, allow a certain degree of flexibility with the accuracy of the transcription itself. 

\citet{amancio2013} conducted an extensive investigation into the statistical properties of unknown texts, using the Voynich Manuscript as a case study. They analyzed the word frequencies within the text, concluding that Voynichese is compatible with natural languages.  \citet{reddy2011} ran a variety of statistical and linguistic tests, and found page-level topics, as well as word and length frequency distributions, that ``conform to natural language-like text.'' Though \citet{rugg2004,rugg2016}, and \citet{Daruka2020} suggest ways in which hoax text could be generated, it is difficult to reconcile the character-level methods they suggest with the document-level structure that \citet{reddy2011} and \citet{amancio2013} recover. That is, while Voynichese appears very un-languagelike at the word level, the Voynich manuscript --- at the level of paragraph and page --- has much in common with natural language texts.

The current paper extends research on word frequencies, focusing on  methods for topic modeling, and  investigates how computer-identified topics relate to clusters of pages and paragraphs  within the manuscript which have been identified on other bases, including illustrations, scribal attribution \citep[e.g.][]{davis}, and the ``languages'' identified by Prescott Currier \citep[e.g.][]{currier1976papers}. 

Here we find evidence for matches between topics and visually identified sections. We also find some support for different topics and different scribes. Finally, we show that scribal hands and subject matter appear to jointly determine topic membership, strongly implying that several linguistic features contribute to the identification of topics.

\section{Background}
\subsection{Computational models and topic analysis}
This section provides information on the background to the methods used. Note in the discussion that follows, a `document' is simply a unit of text. We test several units of text of different sizes, including folios, pages, paragraphs, and 40 word chunks (discussed further below).

\subsubsection{Document Vectorization}
Prior to analyzing the statistical properties of a given corpus, it is necessary to vectorize the documents, i.e. represent the text numerically. In this work, we consider various topic modeling algorithms, which involve two methods for document vectorization: \textit{bag of words} (BoW) and \textit{term frequency-inverse document frequency} (tf-idf). Both BoW and tf-idf are count-based document representations that ignore word order, however tf-idf weights the importance of each word in a document relative to other words in the same document and the entire document collection, whereas BoW only considers raw word counts. In both approaches, a given corpus is represented as a $N \times V$ matrix, where $N$ is the total number of documents and $V$ is the number of words in the vocabulary. 

The tf-idf value for a given word is computed as the product between the term frequency and inverse document frequency:

\begin{equation}
    {w}_{t,d} = count(t,d) \times log_{10} (\frac{N}{df_{t}})
\end{equation}

Here, $count(t,d)$ is the term frequency, or the number of occurrences of term \textit{t} in document \textit{d}, which measures the importance of \textit{t} to the document \citep{luhn1957}. The inverse document frequency, $\frac{N}{df_{t}}$, is the quotient of the total number of documents in a corpus, \textit{N}, and the number of documents in which \textit{t} appears. The intuition of the inverse document frequency, $\frac{N}{df_{t}}$, is that rare terms, i.e. words that appear infrequently in the entire corpus, are important to the document(s) that contain those words \citep{jones72astatistical}. Since there are often a large number of documents in a corpus, a logarithm is applied to reduce the scale of tf-idf values; in other variations of tf-idf, a logarithm is also used to moderate the term frequency. 

\subsubsection{Topic Modeling}
Topic modeling is a technique for identifying relatedness between documents that share sets of words belonging to a particular semantic domain. Topic modeling techniques apply unsupervised learning algorithms to discover `topics', or themes, within documents, and consequently cluster documents based on their content \citep{jurafsky-a}. Although we cannot read the Voynich text, topic modeling is still applicable if we assume that Voynich words have a consistent form--meaning correspondence across the manuscript. That is, we need to assume that \texttt{8ain} on \emph{f1r} is the \textsc{same} word as \texttt{8ain} on \emph{f7v}. We do not need to know what any of the words mean, but we do need to assume that there is some consistency of representation. Note, however, that is there is no consistency of representation, we are unlikely to find clear topic structure in the manuscript. 

In this work, we divide the Voynich Manuscript into chunks of text --- documents --- and apply the following topic modeling algorithms: latent dirichlet allocation (LDA), latent semantic analysis (LSA), and nonnegative matrix factorization (NMF). For LDA, we use the BoW approach to vectorize the documents, and for LSA and NMF, we represent the documents using tf-idf. 

LDA is a generative probabilistic model that explains sets of observations with latent, or unobservable, groups \citep{blei2003a}.  It assumes that documents are produced with a fixed number of words and a mixture of topics that have a Dirichlet distribution over a fixed set of $k$ topics to discover. Each word in the document is generated by picking a topic according to the distribution sampled, and then using that topic to generate a word, according to the topic’s distribution. LDA learns the topic representation of each document, as well as the words associated with each topic. In order to do this, LDA distributes $k$ topics across each word $w$ in document $m$, and for each $w$ in $m$ assumes the topic assignment is correct for every word except the current $w$. Finally, it probabilistically reassigns $w$ a new topic based on the proportion of topics within a document and the proportion of words within a topic. This process is repeated many times, and the model eventually converges, achieving feasible topic mixtures within documents. It is not necessary to weight words using tf-idf for LDA because it's a generative model that estimates probability distributions, thus we use BoW instead.

LSA is another method that takes advantage of implicit higher-order structure in the association of terms with documents to discover underlying concepts within documents \citep{deerwester1990a}. LSA utilizes Singular Value Decomposition (SVD) to reconstruct the matrix such that the strongest relationships are preserved and noise is minimized. SVD follows that any matrix A can be factored as 

\begin{equation}
    A = U S V^{T},
\end{equation} 

where $U$ and $V$ are orthogonal matrices with eigenvectors from $AA^{T}$ and $A^{T}A$, respectively. This process decomposes the matrix into a document-topic matrix and a topic-term matrix, which contain the relative importance of each factor and allow us to measure document similarity. 
         
NMF is used to analyze high-dimensional data and automatically extract sparse and meaningful features from nonnegative vectors \citep{lee1999a}. NMF decomposes the feature matrix $A$ into two lower dimensional matrices, $W$ and $H$, then iteratively modifies their initial values over an objective function (e.g., the EM algorithm) such that their product approaches the original matrix. NMF is similar to LSA and SVD, however NMF imposes the restriction that factored matrices, $W$ and $H$, are positive. As a result, NMF better represents the original feature matrix. 

\subsubsection{Visaulization}
After applying NMF or LSA to the tf-idf document-term matrix, we get a denser and smaller matrix, but still one that is too large to be plotted usefully in two dimensions. We would like to produce a plot that shows folios of the Voynich Manuscript on a 2-dimensional plane, with the distance between points corresponding to the `semantic distance' between the folios that were the output of the algorithms discussed above. This is a general problem in machine learning, as many datasets are multidimensional. Dimension reduction is a difficult problem, and there are an infinite number of ways to project a high-dimensional object into 2D space. Moreover, the choice of dimension-reduction algorithm is important, as the algorithms are biased towards different projections and so the choice of algorithm can strongly influence the result, especially in cases (such as this) where the dataset is very small. 

We consider three different methods for projecting the data:

\paragraph{PCA:} Principal Component Analysis is one of the oldest and best understood dimension reduction algorithms. The first principal component is the axis along which there is the highest variance. The second axis also chooses the highest variance, with the constraint that it must be perpendicular to the first axis. This continues until the required (user-specified) number of axis is satisfied.

\paragraph{t-SNE:} T-Distributed Stochastic Neighbor Embedding is a relatively recent (2008) technique; it examines the local neighborhood around every point, in addition to the properties of the distribution as a whole \citep{van2008visualizing}. This facilitates the detections of small clusters in a large distribution, which might not be picked up by PCA. However, it is more opaque than PCA. t-SNE has been used on vectorizations of the Voynich Manuscript with some success in the past \citep{bunn,perone}. We found that t-SNE was sensitive to the choice of hyperparameters.

\paragraph{UMAP:} Uniform Manifold Approximation and Projection for Dimension Reduction is a newer algorithm \citep{mcinnes2018umap}, which preserves more global structure than t-SNE.


\subsection{Voynich Manuscript background}
The basic facts of the Voynich manuscript are by now clearly summarized in other publications. See, for example, \citet{reddy2011,amancio2013,Bax2014,Harkness2016} and \citet{bowernlindemann20} for a recent overview (book-length treatments can be found in \citealt{kennedy2004} and \citealt{dimperio1978}). Here we focus on three aspects of the manuscript most relevant for document clustering: the ``Languages'' of the text, number of scribal hands, and subject matter as inferred from the manuscript's illustrations.

\subsubsection{Currier Languages}
\citet{currier1976papers} observed  differences in character and substring frequencies across different sections of the text. Consequently, he split the text into ``languages,'' or dialects A and B, and analyzed the two languages separately.\footnote{The original Currier designation included several pages that could not be assigned to either of the main hands. Note also that while Currier discussed multiple languages, discussion has tended to focus on just two (called ``A'' and ``B'' for convenience); we follow that tradition here.} From there, he discovered that symbol groups appearing very frequently in one language may be almost non-existent in the other. From a statistical perspective, this discovery has complicated our understanding of the Voynich Manuscript because topic modeling relies on word frequencies and expects consistency across texts; for this reason, we consider the topic distributions in conjunction with Currier languages. Other Voynich investigations have tended to focus on a single part of the manuscript. For example, \citet{reddy2011} use only data from the B language for their investigations. Note that some discussion of Currier languages and further topics can be found in \citet{lindemanbowern20}.

\subsubsection{Manuscript illustrations}
On the basis of illustrations which accompany the text, it is customary to divide the manuscript into five sections.

\begin{enumerate}\addtolength{\baselineskip}{-2pt}
\item  botanical/herbal
\item astrological/astronomical
\item  balneological
\item  pharmaceutical
\item  starred paragraphs/“recipes”
\end{enumerate}

Here, we use a larger set of divisions, also including the large foldout 9-sectioned rosette diagram as its own `subject', and an `unknown' section for the few pages which comprise only text and cannot be aligned with other subjects on the basis of illustrations. 

\subsubsection{Lisa Fagin Davis’s Five Scribe Theory}
\citet{davis} proposes that the manuscript was written by five scribes. Her approach involves identifying paleographic features (that is, glyph shapes) that distinguish scribes from each other.  Likewise, these hand attributions complicate our understanding of Voynichese as a language, prompting us to consider whether the text contains different linguistic dialects, or if written variations of the same word may be transcribed as different words. Alternatively, perhaps different scribes were responsible for different sections (or subjects), which would probably be detectable with topic modeling based on word frequency. Thus, we apply statistical analysis assuming Davis' identification of hands in the manuscript.

\begin{figure*}[ht]
    \centering
    \includegraphics[width=\linewidth]{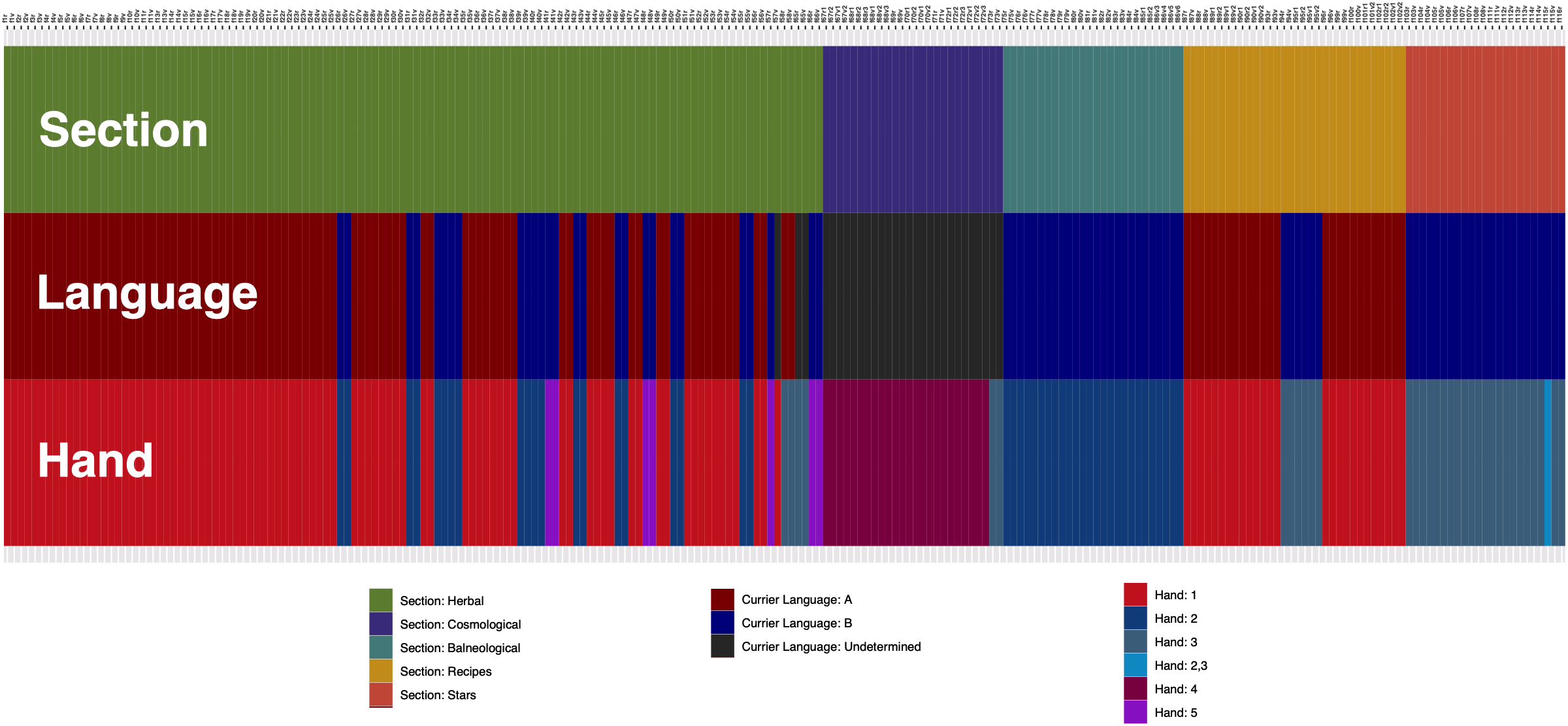}
    \caption{Alignment of languages, illustrative subjects (here labeled `section'), and hands in the Voynich Manuscript, reproduced from \citet{lindemanbowern20}}
    \label{fig:vmap}
\end{figure*}

Figure~\ref{fig:vmap}, reproduced here from \citet{lindemanbowern20}, illustrates how the scribes, subjects, and Currier languages align throughout the manuscript. As can be seen, there is some overlap in assignment to languages and section: for example, while the botanical section has folios in both A and B languages, the balneological and stars sections are in language B alone. Language and hand also aligns well, though there are more `hands' than `languages'.

\section{Data and Methods}
The Voynich transcript used here is Takahashi's version of the text (as corrected by Zandbergen and Stolti) in the EVA transcription system. The Takehashi transcription is complete and well-regarded, but does require some processing before it can be used. We used a ‘tokenizer’ script to remove unnecessary annotations and characters from the transcription. The Takehashi transcription marks the location of breaks in the text for illustrations, as well as some other information about the location of the text on the page. The Tokenizer removes all of this, separates each word out as a single string, and groups them by page. The vectorization and clustering algorithms receive this and only this as input. No information is conveyed to our analysis tools about the formatting of the manuscript or illustrations on a page.

The text was  vectorized into word count matrices representing the text. Additionally, Davis’s 5 scribal attributions were used to  construct a dataset of possible internal sections, along with language classification (language A, B), visual subject label (botanical, astrology, balneological, rosette, recipes, starred paragraphs, and unknown), and quire number (1-18). For each topic modeling analysis,  the topic classifications generated by the algorithm were appended and compared.\footnote{Folio f57v was excluded from the analyses because its Currier language is unknown.}

\subsection{Vectorization}
Given that there is no translation of Voynich text, it is impossible to determine which words are most meaningful, or which share similar meaning, by reading the manuscript. However, we can work out which words are most characteristic of which parts of the text. The first step of our process is to vectorize the document. This converts each word and each document into a vector in a very high dimensional space. Vectorization allows us to compute information about words, documents, and relationships between them, even with no understanding of the words or documents themselves. Once the folios and words have been vectorized, it begins to make sense to talk about how ‘close’ two folios are to each other. If the Voynich Manuscript were gibberish, we would expect the ‘distances’ between folios to be essentially random. However, if the VMS contains natural language with genuine meaning, we would expect to see that two pages that seem to cover the same topic (e.g. two herbal pages) would be closer to each other than to pages that seem to be about other topics. Note that there are several types of similarity which might be detected by this method: either meaning, type of encipherment, or perhaps a combination of both.

\subsection{Clustering}
With these vectors in hand, the next step is to systematically locate clusters of vectors that are close to each other. In general, clustering aims to find structure in a set of vectors by labeling nearby vectors as a single object.

\begin{table}[ht]
\begin{small}
\begin{tabular}{p{.4\linewidth}p{.4\linewidth}}
\hline
 Clustering Inputs & Clustering Outputs \\ \hline
 Vectorized document corpus & Cluster ‘locations’\\
 Number of clusters & ‘Distances’ from each vector to each cluster\\
 \hline
 \end{tabular}
 \end{small}
 \centering\caption{Clustering inputs and outputs}\label{tab:cl}
 \end{table}
 
We tell our algorithm how many clusters to make (that is, the value of $k$), but not what those clusters are. The clustering algorithm then determines where to place the clusters, and from those locations we can determine the nearest cluster to each folio, categorizing them into ‘topic’ clusters.

\subsection{Dimension reduction and visualization}
At this point in our process, we have vector forms of Voynichese words and VMS folios, as well as a ‘cluster’ label for each folio. However, these vectors are in a high-dimensional space, and are thus impossible to directly plot or visualize. We would like to produce a 2D plot depicting each folio as a point in our vector space. The choice of which two dimensions to use is not trivial -- even simple objects can look different if they are rotated, and our distribution has far more potential rotation axes than a real 3D object. In addition, simply selecting two dimensions of the current coordinate system is unlikely to be sufficient, as it would loose all of the information in the many other dimensions. Instead, a complicated projection is needed, that can collapse the data to two dimensions, while maintaining characteristic overall structures.

Fortunately, this is a common problem in machine learning, because high-dimensional vector spaces are a generally useful way to work with a dataset, and one often wants to plot multidimensional data in two dimensions eventually. For this project, we made use of three established algorithms for dimension reduction: PCA, t-SNE, and UMAP. 

As discussed above, three topic modeling algorithms were tested: latent dirichlet allocation (LDA), latent semantic analysis (LSA), and nonnegative matrix factorization (NMF). LDA uses raw word frequencies, whereas LSA and NMF use tf-idf weighted counts. Finally, we used multiple correspondence analysis (MCA) to compare the topic modeling results to qualitative features of the text, including hand attribution, Currier language, and illustrative topic classification.

\subsubsection{MCA}
For each of these algorithms, we used MCA (Multiple Components Analysis), an extension of correspondence analysis used to analyze the relationship of categorical dependent variables \citep{abdi2007a}. By representing data points in a 2-dimensional space, MCA visualizes the relationship between the topics generated by the topic modeling algorithms to qualitative measures of hand, Currier language, and illustrative topics. MCA is a dimension reduction technique that uses an indicator matrix, which contains rows of individual data points, columns representing variable categories, and entries of 0 or 1. Relationships between variables are discovered by computing the chi-square distance between categories of variables and individual data points. MCA discovers the underlying dimensions that best explain differences in data, which allows for the data to be represented in a reduced space. 



\section{Results}
We report on all results, informative and uninformative. Informative results are summarized in the conclusions but we felt it was important to also show the approaches that failed, and to discuss why.

\subsection{Analysis 1: Latent Dirichlet Allocation}
\begin{figure}[ht]
    \centering
    \includegraphics[width=\linewidth]{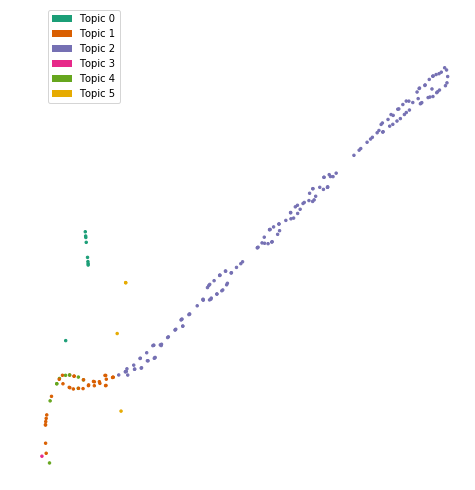}
    \caption{LDA 6 topic distribution}
    \label{fig:LDA}
\end{figure}

The results of LDA topic clustering are shown in Figure~\ref{fig:LDA}. 

\begin{figure*}[ht]
    \centering
    \includegraphics[width=.9\linewidth]{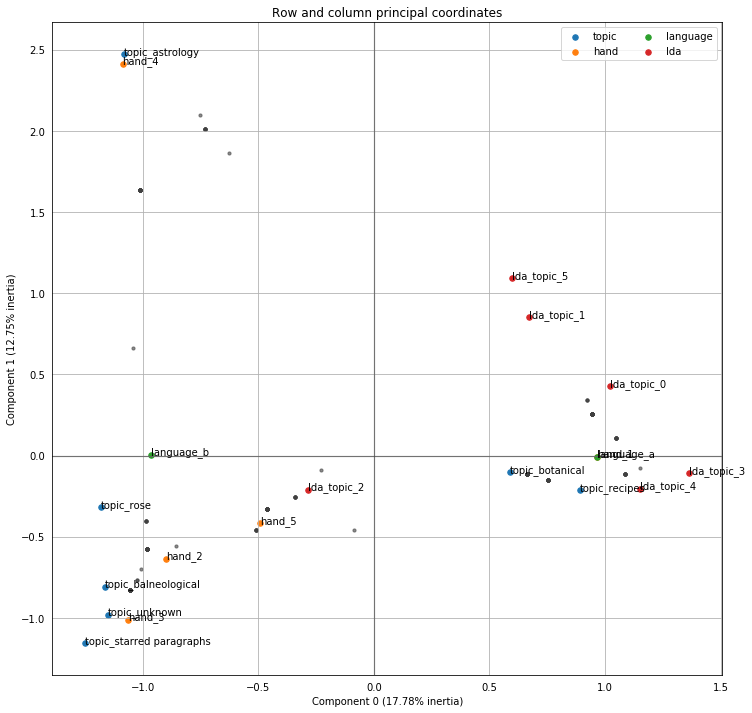}
    \caption{MCA for LDA topics, illustrative topics, languages, and hands}
    \label{fig:MCA-LDA}
\end{figure*}

The LDA model created one dominant topic, with the majority of pages belonging to topic 2. The distribution of scribes, languages, visual topics, and LDA topics (visualized with MCA) are shown in Figure~\ref{fig:MCA-LDA}. 

The MCA results to visualize the LDA clusters are not informative, as the LDA topics don’t seem to cluster in any significant way. However, it is interesting to note that there is a clear distinction between Language A and B in the results. The top words per topic for LDA are shown in Appendix~\ref{AppA}.

While it may be possible that the manuscript is mostly about one topic, there are certain aspects of LDA that may explain these results and render LDA unfeasible to use for these analyses. LDA expects a good understanding of the text data, as it groups co-occurring words together. When performed with optimal parameter settings, the topics generated by LDA are close to human understanding. If topics share keywords, smaller topics can be absorbed into a major one, which usually indicates a suboptimal parameter setting \citep{ma2016a}. This seems to suggest several reasons why the LDA analysis is not accurate: perhaps there are insufficiently clear word clusters (that is, that the topics are not sufficiently lexically differentiated to show up by this method), or, conversely, that spelling variation obscures lexical identity. Note, however, that \citet[83]{reddy2011} were able to find clusters (but they provided no information about their approach beyond the use of tf-idf). It is also possible that the Voynich manuscript isn’t human language, or the Takahashi tokenizer isn’t sufficiently accurate. It is typically encouraged to understand the structure of data using other methods before using LDA, which is why we use NMF in the following analyses (Analyses 4--6). 

\subsection{Analysis 2: LSA}
\begin{figure*}[ht]
    \centering
    \includegraphics[width=.9\linewidth]{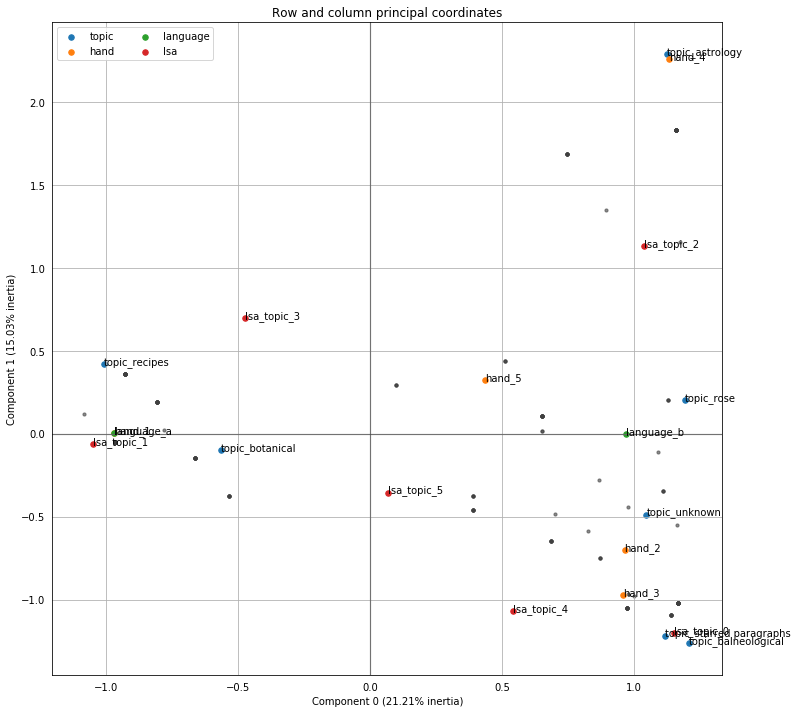}
    \caption{MCA for LSA topics, illustrative topics, languages, and hands} 
    \label{fig:MCA-LSA}
\end{figure*}

\begin{figure}[ht]
    \centering
    \includegraphics[width=\linewidth]{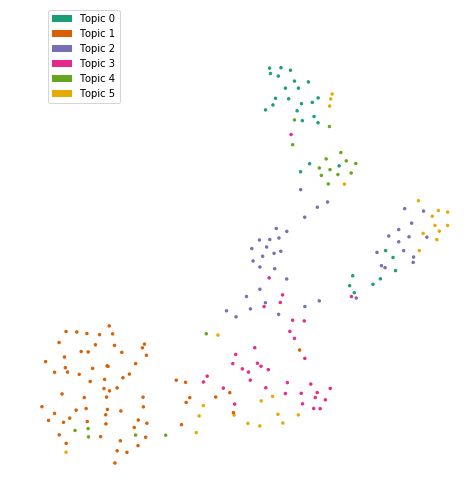}
    \caption{LSA 6 topic distribution}
    \label{fig:LSA}
\end{figure}

The results of LSA topic clustering are shown in Figure~\ref{fig:LSA}, and the distribution of scribes, languages, visual topics, and LSA topics from MCA are shown in Figure~\ref{fig:MCA-LSA}.

The clusters for LSA are more distinct than those of LDA. For both LSA and LDA, hand 4 and the astrology visual topic are quite close to each other. This is expected, since Hand 4 is responsible for this section of the MS (this is a check that sensible results are returned). 

Again, we see a distinct contrast between Language A and B, though it is perhaps not as strong. In this LSA analysis, Language A and hand 1 are overlapping in the figure, with LSA topic 1 very close by. Additionally, LSA topic 0 and the starred paragraphs illustrative topic are almost identical, with balneological illustrative topic very close by.\footnote{It’s worth noting that LSA assumes a Gaussian distribution, however words in documents may follow a Poisson distribution.} 

Additionally, LSA does not perform well on texts with large amounts of polysemy or homophony, since it assumes words only have one concept. We do not know if this is relevant to the Voynich Manuscript, but if the cipher results in the conflation of phonemic distinctions (implied by H2 entropy, per \citealt{lindemanbowern20}), it is quite possible that there are more Voynich homophones than in the underlying language.  While future research into LSA may prove useful, we used NMF as our primary topic modeling algorithm for the remaining analyses, given the issues identified above. The top words per topic for LSA are shown in Appendix~\ref{AppB}.

\subsection{Analysis 3: NMF}
\begin{figure}[ht]
    \centering
    \includegraphics[width=\linewidth]{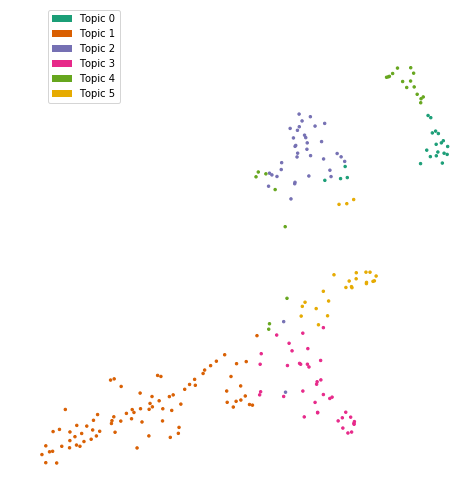}
    \caption{NMF 6 topic distribution}
    \label{fig:NMF}
\end{figure}

The results of NMF topic clustering are shown in Figure~\ref{fig:NMF}. There is a clear divide between the clusters representing Language A and B along the horizontal axis.  This distinction is much stronger than those found in the analyses using either LDA or LSA. 

\begin{figure*}[ht]
    \centering
    \includegraphics[width=.9\linewidth]{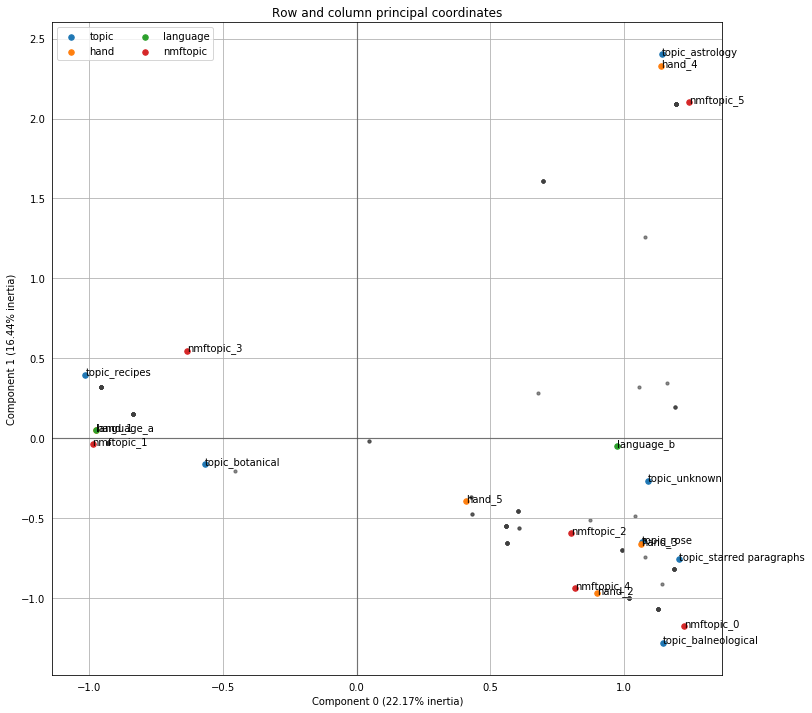}
    \caption{MCA for NMF topics, illustrative topics, languages, and hands} 
    \label{fig:MCA-NMF}
\end{figure*}

The distribution of scribes, languages, visual topics, and NMF topics are shown in Figure~\ref{fig:MCA-NMF}.

The MCA for NMF shows closer relationships between NMF topics and the illustrative sections. As with the case of both LSA and LDA, astrology is clustered with hand 4. However NMF differs from the two previous analyses because astrology and hand 4 are also shown to be related together NMF topic 5. That is, we get an overlap between the hand, the manuscript section (as indicated from the illustrations), and the topic identified through NMF. Note that the computational analysis has  information about neither scribe nor illustration, so aligning all three  is unlikely to occur by chance.
  
Illustrative topics of recipes and botanicals are distinct from the other topics, and both are quite close to language A, hand 1, NMF topic 1, and NMF topic 3. This may suggest that recipes and botanicals are distinct topics with overlapping content. The same could be said for illustrative sections of rosette diagram, balneological material, and the starred paragraphs. The balneological section is very closely related to NMF topic 0, and the rosette and starred paragraphs are both quite close to NMF topics 2 and 4. Taking a closer look at the hands, we see that hand 2 and NMF topic 4 are extremely close, hand 3 and illustrative rosette topic are almost identical. The top words per topic for NMF are shown in Appendix~\ref{AppC}.

It’s interesting to observe that the NMF topics have close relationships with both illustrative sections and hands; this supports the notion that NMF topic does not directly correspond to either hand or illustrative topic---rather that features of both may be reflected in NMF topics. What these features are precisely, are not detectable through these methods, but they must relate to aspects of word use. That could be spelling, encipherment patterns, or word choice, for example. We do not speculate further at this point on \emph{why} hand shows up as a factor that is distinct from illustrative section.

The results of NMF clustering showed closer relationships with hands, illustrative topics, and Currier languages, than those of LSA and LDA. Unlike LDA, NMF performs better with initial data explorations because it's better at handling noise \citep{ma2016a}. In the case of Voynich Manuscript, this noise may be introduced by the Takahashi transcription; there is no way, of course, to verify that the transcription accurately divides the text into linguistic units, such as words. Of course, the Voynich dataset is ``noisy'' on several dimensions: in the digitization of the transcription and the small amount of data, to name just two factors. LDA performs better when the data reflects ``semantic units'', whereas NMF performs better with unstructured data than the other methods do. For these reasons, we consider NMF to be the most reliable algorithm for topic modeling the Voynich Manuscript, and will continue to use NMF for Analyses 4--7, as further detailed below.

\subsection{Analysis 4: Fixed document lengths with random word selection}
In previous analyses, the unit of analysis was the `page'. However, pages have very different amounts of text. Consider the amount of text on a star chart, compared to one of the pages of starred paragraphs, for example. It is possible that the length of the page may be distorting similarity.\enlargethispage*{-1cm} 

In order to eliminate bias introduced by varying document lengths, analyses 4a and 4b use subsets of the Voynich data by randomly selecting 40 and 20 words, respectively, from each page. Accordingly, it excludes the astrological pages (f67-f73), as the text is mostly labels, and pages with fewer than 50 words,\footnote{These are f5v, f11v, f25r, f38r, f65r, f65v, f90r2} each of which belong to the botanical visual topic. An alternative analysis would be to take the first 40 words of each page, or to split the pages into sub-documents (this, however, would over-represent pages with more words).

\subsubsection{4a: Document length of 40 words}
\begin{figure}[ht]
    \centering
    \includegraphics[width=\linewidth]{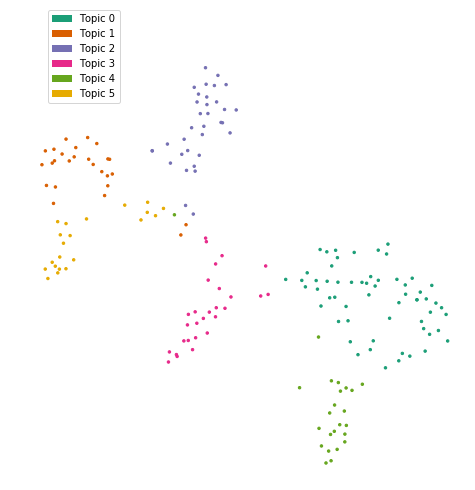}
    \caption{NMF topic distribution with 40 word documents}
    \label{fig:NMF40}
\end{figure}

\begin{figure*}[ht]
    \centering
    \includegraphics[width=.85\linewidth]{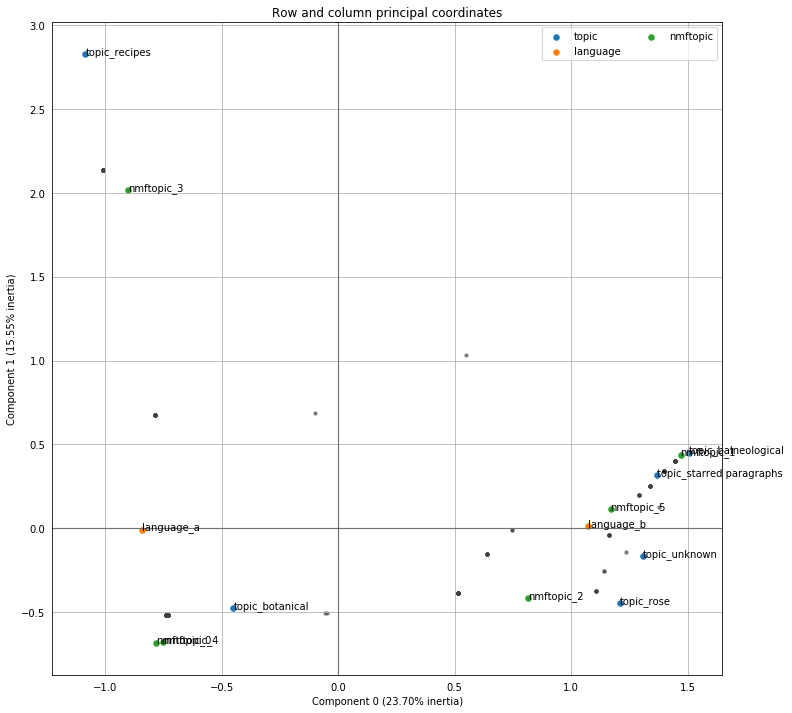}
    \caption{MCA for NMF topics, illustrative topics, languages, and hands, using a 40 word window} 
    \label{fig:MCA-NMF40}
\end{figure*}

As shown in Figure~\ref{fig:NMF40}, documents of 40 random words still maintain strong clusters. Furthermore, Figure~\ref{fig:MCA-NMF40} shows clear distinctions between language A and B.

The MCA reflects a close relationship between NMF topic 1 and the balneological illustrative topic. The star paragraphs are close to NMF topic 1 and the visual balneological topic, but it also is close to NMF topic 6.\footnote{Remember that topic numbers are arbitrary.} Again, this suggests that the starred paragraphs and balneological may have overlapping content as distinct topics. 

One striking difference between the page-level NMF analysis and this analysis with a 40 word cap is that the illustrative botanical and recipe sections are distinct both from each other and from every other section (as defined by illustrations). It’s worth noting again that the folios removed for being less than 50 words were all botanical; this perhaps suggests those seven folios provide some link, in the form of overlapping words, between recipes and botanicals. Three of the pages (\emph{f65r, f65v}, and \emph{f90r2}) are botanical pages which occur in the manuscript adjacent to or among the recipes. \emph{f65r,v} are immediately preceded by missing pages, and \emph{f90r2} is one of a set of herbal illustrations within the recipes section.

Additionally, the recipes section is not very closely related to NMF topic 3, however the two points are much closer with each other than any other point in the figure. NMF topics 0 and 4 are almost identical with the botanical section close by. It’s interesting to see NMF topics 0 and 4 nearly collapse into one topic; this may suggest that they vary on some dimensions or features not represented by the figure, or that they are so similar that they may be considered one topic.

\subsubsection{Analysis 4b: Document length of 20 words}
\begin{figure}
    \centering
    \includegraphics[width=.9\linewidth]{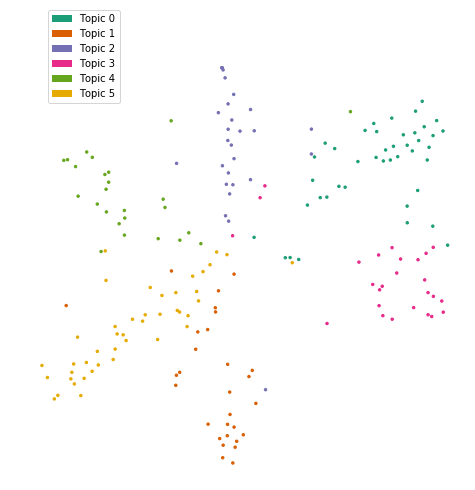}
    \caption{NMF topic distribution with 20 word documents}
    \label{fig:NMF20}
\end{figure}

As shown in Figure~\ref{fig:NMF20}, reducing the document window to 20 words still allows for topic clustering. However the clusters are not as distinct from each other as they are with 40 words or with the full text. 

\begin{figure*}[ht]
    \centering
    \includegraphics[width=.8\linewidth]{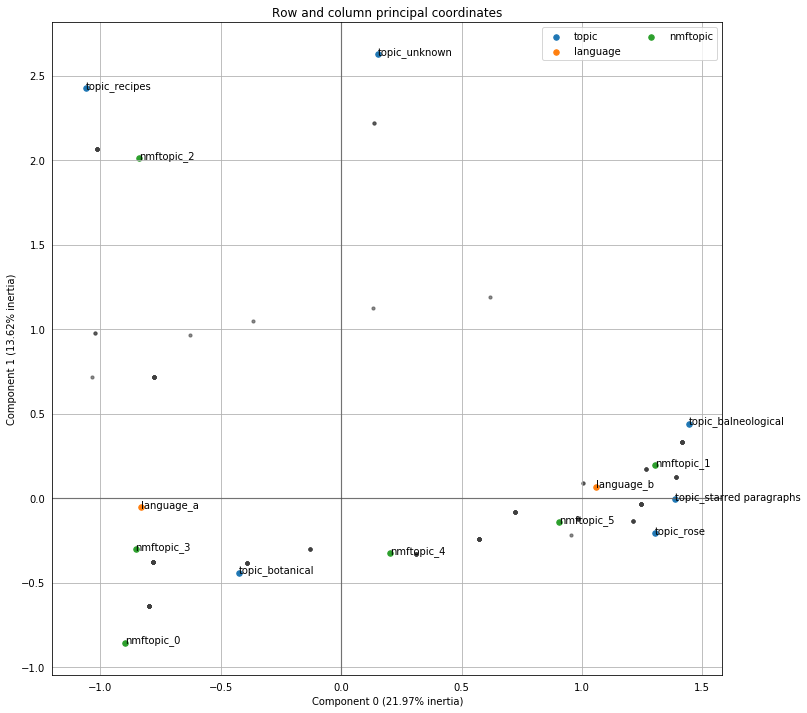}
    \caption{MCA for NMF topics, illustrative topics, languages, and hands, using a 20 word window} 
    \label{fig:MCA-NMF20}
\end{figure*}

Likewise, Figure~\ref{fig:MCA-NMF20} also shows that the language A and B distinction is not revealed as strongly. One stark difference from 40 words to 20 words is that the ``unknown'' visual topic becomes its own cluster, distinct from any other feature, though closer to the recipes section than the balneological, starred, and rosette sections. It’s also interesting to see that the starred paragraphs fall somewhere between the rosette and balneological visual topics, whereas it was much closer to the balneological visual topic for analyses with the 40 word window, and much closer to the rosette for the full text NMF analysis. 

It is also intriguing to see that the structure of the overall cluster of a 20 word NMF analysis is quite similar to that of the full text (page-level) NMF analysis. This suggests that the words in each of the topics are distinct enough from each other that only a small portion of randomly selected words from the text are enough to distinguish documents from each other. Such a result strongly implies that the topics discussed here are not artefacts of analysis on a manuscript with ``gibberish'', and that there is, in fact, meaningful content underlying an enciphered text. However, the fact that there is a difference between 20 words and 40 suggests that words in the Voynich manuscript are a combination of distinctive content words and less distinct function words (or content words that do not uniquely define the topic under discussion). This again suggests that the underlying text is not gibberish, or the result of locally random fluctuations in ``word'' production: if that were the case, and a different type of non-word production was used for each page, we would expect any set of methods that work at the page level to perform roughly equally well.

\begin{figure}
    \centering
    \includegraphics[width=\linewidth]{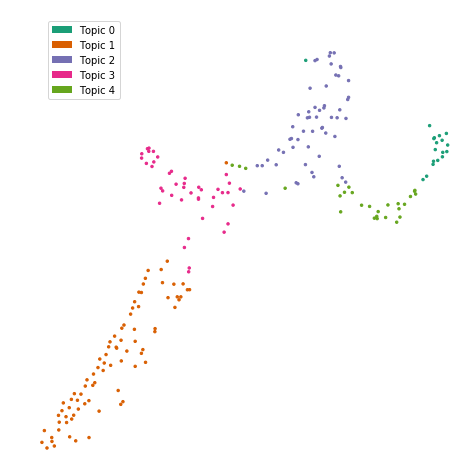}
    \caption{NMF 5 topic distribution}
    \label{fig:NMF5top}
\end{figure} 
\subsection{Analysis 5: NMF topics vs. scribe hands}

 While topic modeling provides a “semantic” understanding of the text, it is still difficult to infer what linguistic features are reflected by topics. Comparing the NMF topics to Lisa Fagin Davis’s hand attributions may offer more insight into what a topic is. For these analyses, we apply NMF with 5 topics to explore whether there is a direct mapping between 5 NMF topics and the 5 scribes identified in \citet{davis}. 

\begin{figure*}[ht]
    \centering
    \includegraphics[width=.85\linewidth]{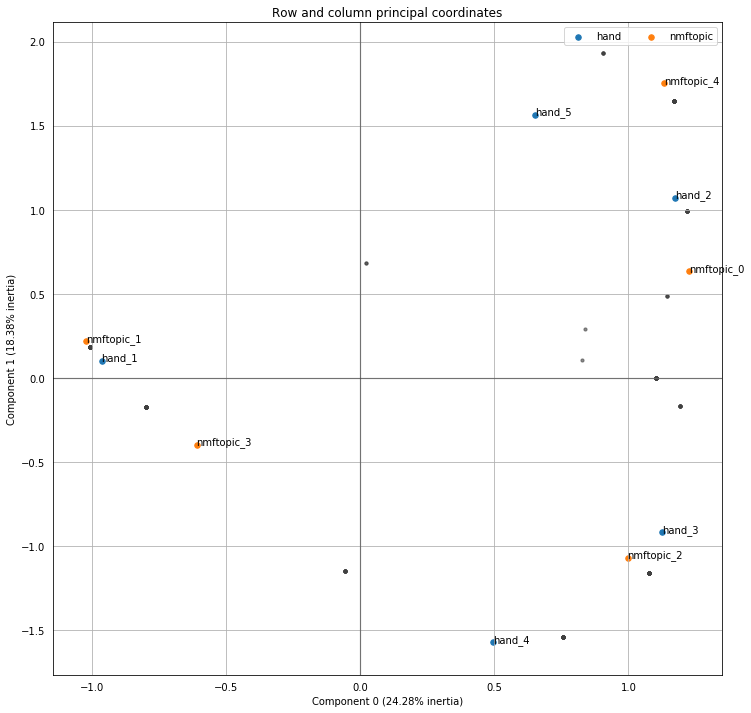}
    \caption{MCA for NMF 5 topics and hands} 
    \label{fig:MCA-NMF5top}
\end{figure*} 

As shown in Figure~\ref{fig:NMF5top}, most of the folios fall into topic 1. Likewise, Figure~\ref{fig:MCA-NMF5top} finds that NMF topic 1 and hand 1 are almost identical. Additionally, NMF topic 2 and hand 3 show close similarity. Hand 2 is somewhat close to NMF topics 0 and 4. 

It is also interesting to note the spread of the clusters: there seems to be one larger cluster on the left side of the graph, one larger cluster in the top right corner, and one larger cluster in the bottom right corner. This division into three larger clusters may have implications for semantic content or scribal similarity. 

Overall, the results suggest that “topic” as defined by NMF is not quite synonymous with hands. We cannot create a complete one-to-one mapping between NMF topics and Voynich scribes; in other words, perhaps not every hand wrote about unique topics.

\subsection{Analysis 6: NMF topics vs. Currier languages}
Scholars often classify sections of the text as belonging to Currier A or B. Thus, this analysis seeks to understand if the Currier languages are reflected by a two topic NMF analysis.\enlargethispage*{1cm}

 \begin{figure}
    \centering
    \includegraphics[width=.8\linewidth]{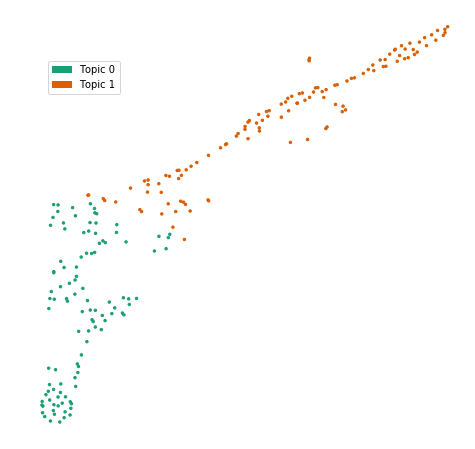}
    \caption{NMF 2 topic distribution}
    \label{fig:NMF2top}
\end{figure}  
 \begin{figure}[h]
    \centering
    \includegraphics[width=.75\linewidth]{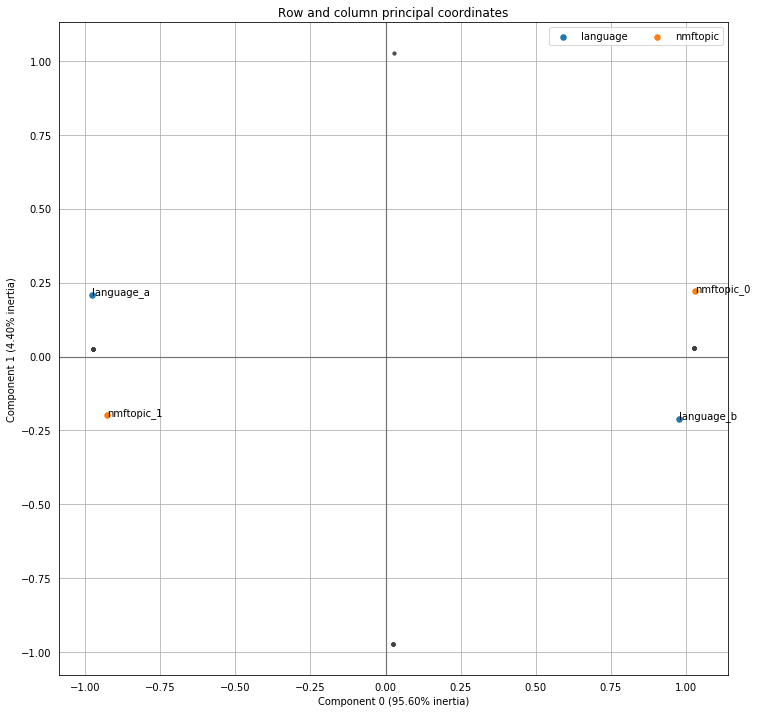}
    \caption{MCA for NMF 2 topics and languages}
    \label{fig:MCA-NMF2top}
\end{figure} 

As shown by Figure~\ref{fig:NMF2top}, the text has distinctive clusters between the two topics. Interestingly, Figure~\ref{fig:MCA-NMF2top} shows that while the Currier languages may share features with the topics, they are not the same; that is, two topics identified by tf-idf do not overlap with the languages identified according to Currier. It’s also worth noting that languages A and B were developed mostly based on character and substring frequencies, whereas NMF topics are based on word frequencies. Thus, the basis of this analysis may be comparing two different  units (Currier symbol groups vs.\ Takahashi words). 


\subsection{Exploring the relationship between hands, topics, and illustrations}

In the previous analyses, we showed how TF-IDF topics do not clearly match to either Hands or Illustrative sections, though there is some overlap in association with both (not surprisingly, since there is an association between scribe and illustrated section). In this section, we further explore the ``network'' links between the TF-IDF topics and other divisions in the manuscript. To do so, we use a network as a visualization device, where nodes in the network are the categories under consideration (hand, illustrated sections, TF-IDF topics, etc), and the edges are the pages the link the hands to sections or topics.

\begin{figure}[h]
    \centering
    \includegraphics[width=\linewidth]{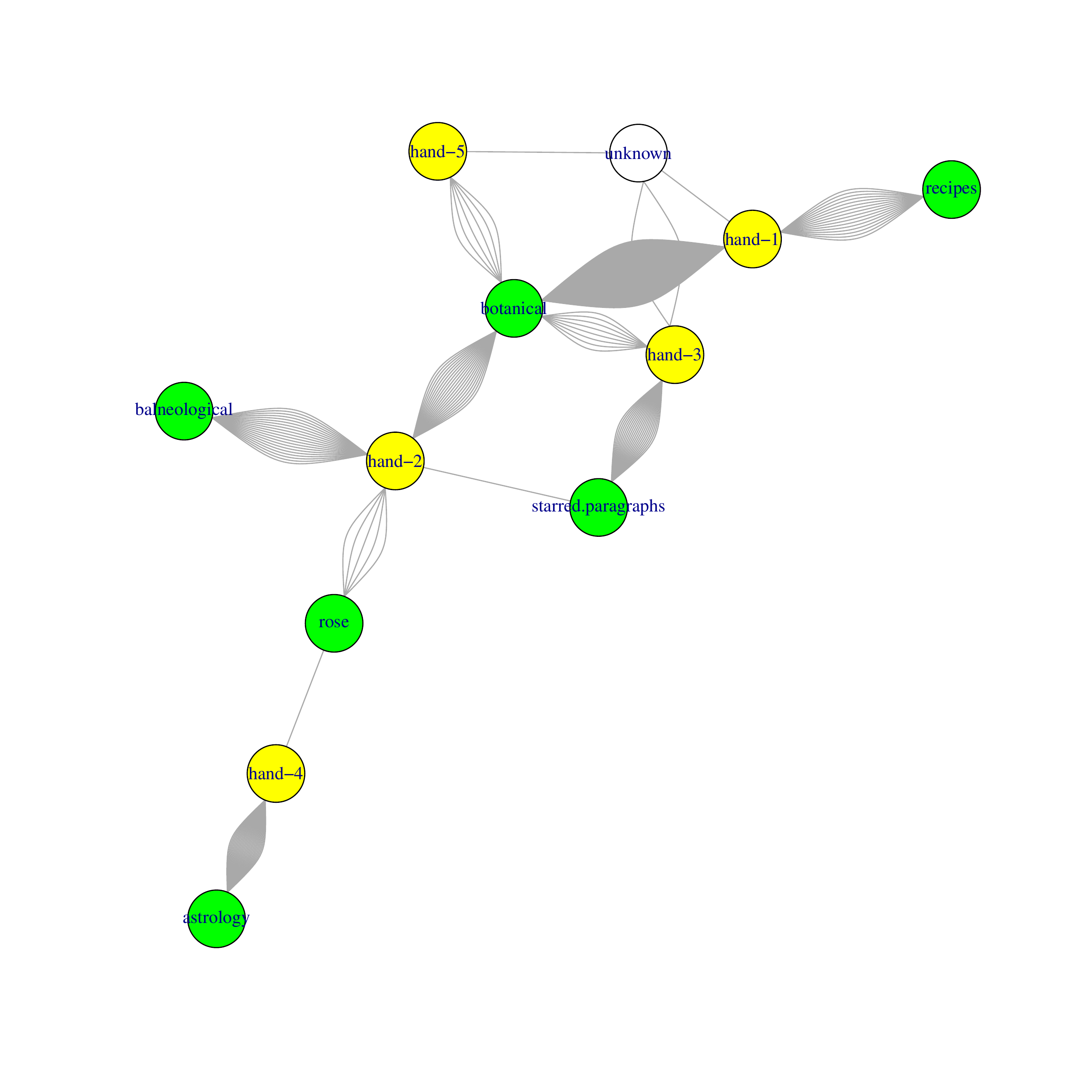}
    \caption{Manuscript hands and illustrated subjects}
    \label{fig:HSnet}
\end{figure}

Consider Figure~\ref{fig:HSnet}. In this network, each node is a hand (yellow) or a subject (green), and the links between them are pages. The assignments are from \citet{davis}. From this, we can see that Hand 4 is associated with the astrology section; Hand 3 mostly wrote the starred paragraphs and botanical pages; Hand 1 wrote the recipes section and also collaborated (with Hands 2 and 3) on the botanical pages, and so forth.

\begin{figure}[ht]
    \centering
    \includegraphics[width=\linewidth]{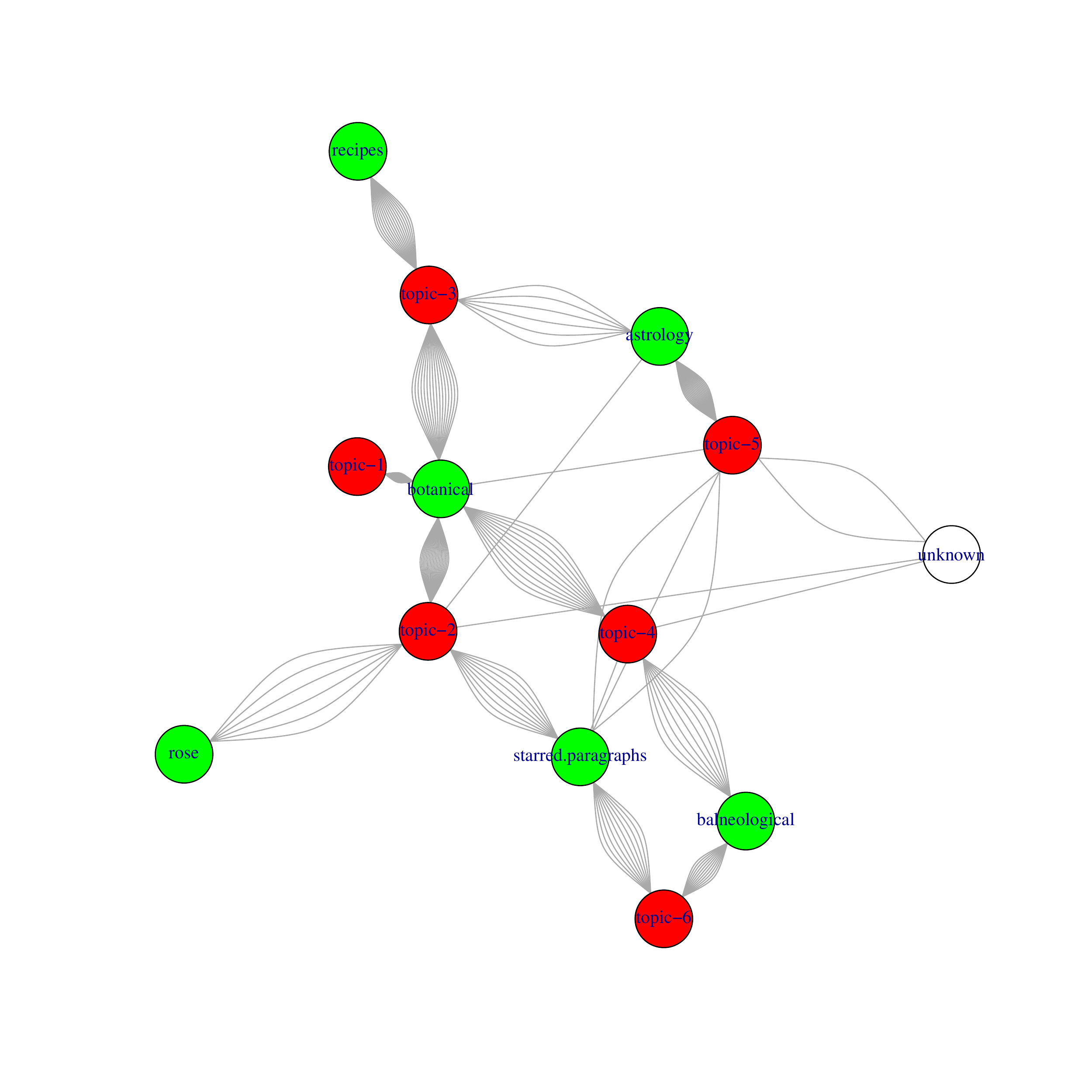}
    \caption{tf-idf generated topics matching illustrated subjects}
    \label{fig:TSnet}
\end{figure}

This same visualization technique can be used to combine illustrated subjects and tf-idf topics. This is shown in Figure~\ref{fig:TSnet}, with topics in red (as identified by tf-idf NMF 6 topics using full pages) and subjects in green. Just as with the scribes/hands, we see that tf-idf topics do not uniquely correspond to the sections as defined by illustrations. The balneological section, for example, is split between topics 4 and 6; the starred paragraphs mostly between topics 2 and 6 (but also with one page assigned to topic 4 and 3 to topic 5).

\begin{figure}[ht]
    \includegraphics[width=\linewidth]{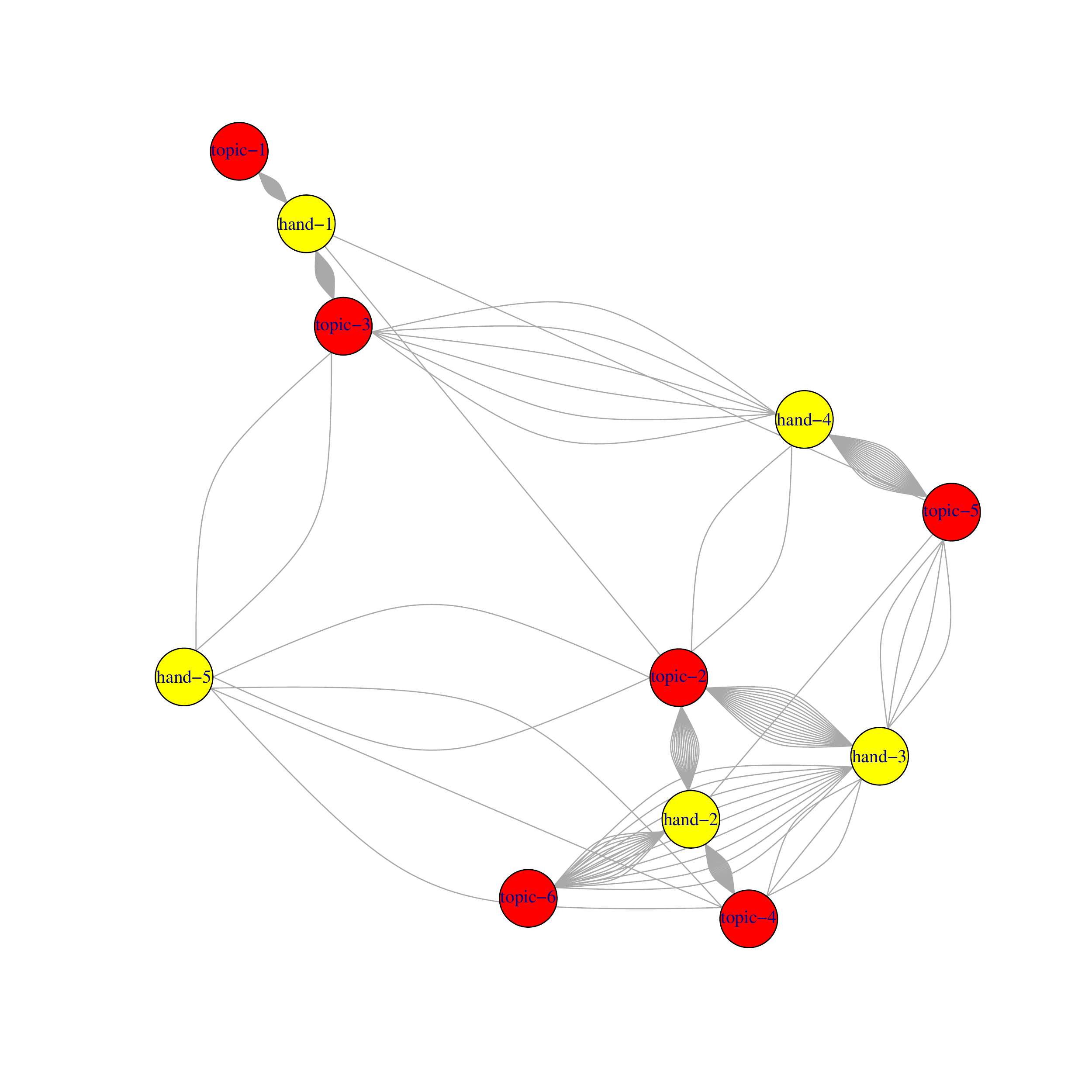}
    \caption{tf-idf generated topics matching scribal hands}
    \label{fig:THnet}
\end{figure}

We can look at Hands and topics, as shown in Figure~\ref{fig:THnet}. Here, the patterns are quite clear: Hand 1 wrote topic 1, and collaborated with Hand 4 (and Hand 5) on topic 3; Hand 4 collaborated on topics 3 and 5, with Hand 1 and 2 each contributing one page to topic 5 as well. Hand two contributed to topics 2, 4, and 6; they were the majority author of topic 4 (though with input from Hand 3) and topic 6 (again with input from Hand 3) while toic 2 is equally split between Hands 2 and 3.

\begin{figure*}[ht]
    \centering
    \includegraphics[width=.9\linewidth]{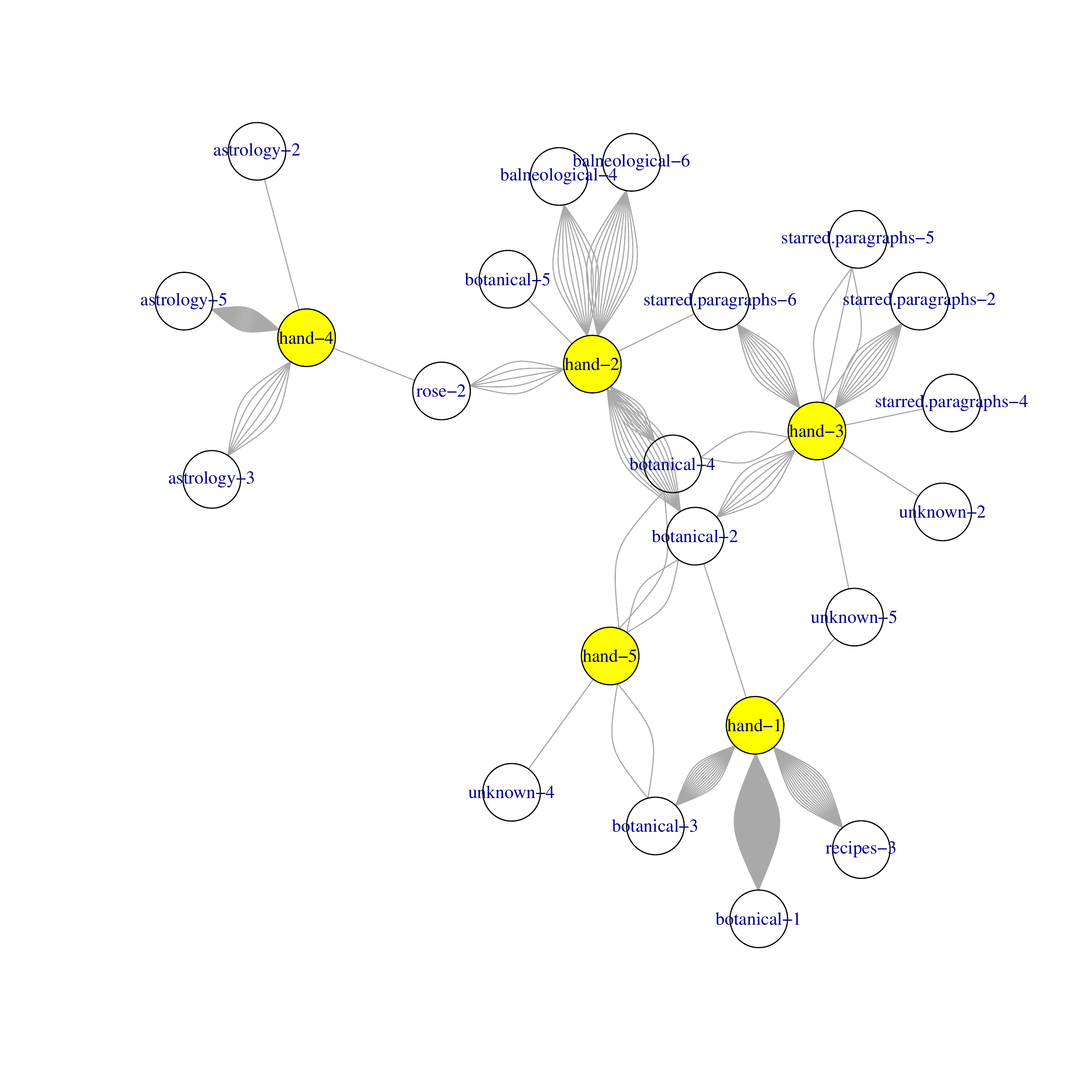}
    \caption{Topic: tf-idf generated topics matching manuscript hands and illustrated subjects}
    \label{fig:tfidf}
\end{figure*}

Finally, we can combine the hands, topics, and subjects, to see if there is further structure and whether we can see anything further about the topic collaboration. This is illustrated in Figure~\ref{fig:tfidf}. In this figure, the yellow nodes are the hands, and the white nodes are a combination of illustrated subject and tf-idf topic. The figure is somewhat difficult to read. Consider the astrology sections, which are divided between topics 2, 3, and 4 in the tf-idf. Here, astrology-2 and -3 are from scribe 4, while astrology 4 is by hand 3. That is, the parts of the astrology section attributed to tf-idf topic 4 is written in  hand 3. Other parts of topic 4 are also written by hand 3: the starred paragraphs in topic 4 and the unknown/unassigned illustrated section. 

When combining tf-idf topics and illustrated sections, one can see further structure associated with the tf-idf topics and hands. In fact, it is striking how well this combination of illustrations and computationally-derived topics divides the manuscript. Botanical-3 is mostly by Hand 1, except for two pages assigned to Hand 5. Hand 2 is responsible for the balneological topics 4 and 6 and botanical 2, 4, and 5 (that is, the parts of the botanical section in topics 2, 4, and 5). Hands 2 and 3 collaborate on botanical topic 2 (and hand 3 does two pages of botanical 4). Apart from the pages associated with hand 5, there are clear associations between hands and tf-idf topics, as well as tf-idf and illustrated sections.  For example, Hand 1 writes the recipes (associated with tf-idf topic 3) and is the sole contributor to botanical topic 1; they are also associated with botanical topic 3. Hand 3 is associated with botical-2 and starred paragraph-2 (that is, the parts of those illustrated sections which are assigned to the same tf-idf topic). The correspondences are not perfect, but they are considerably better than chance.


\section{Further Discussion}
Analyses 1--4 consider the relationships between categorical dependent variables, whereas analysis 5 isolates NMF topic and scribe, and analysis 6 isolates NMF topic and Currier language. These analyses provide strong evidence that scribe, NMF topic, and Currier language are not the same. This, perhaps, provides a more convincing argument that NMF reflects distinct semantic and linguistic features, potentially also including differences in scribe and/or dialect. 

These topic modeling analyses may also offer evidence that some of the texts are written about the illustrations that accompany them. Although the astrological section has fewer words than the other folios and appears to be mostly labels, the NMF full text analysis (Analysis 3) considers the illustrative astrological section its own topic, and its distance is quite far from the other NMF and illustrative topics. Furthermore, in Analyses 1--3, the astrological section is always clustered next to hand 4, although it is worth noting that both hand and illustrative topic classifications come from Lisa Fagin Davis’s work. That is, the topic is associated with that hand, even though no information about the hand was used in discerning the topic.

It’s also interesting to see the close clustering of illustrative rosette, balneological, and starred paragraphs topics across the analyses. The analyses seem to suggest that these sections belong to a larger common topic, though there are individual differences. This is particularly interesting because the starred paragraphs and parts of the rosette sections offer few visual clues as to what the text may be about, whereas a topic modeling approach --- even with a fixed count and random word selection --- suggests they may be related to the balneological illustrations and text.  Likewise, the increase in distance between the recipes and botanical visual topics after several botanical pages were removed from the analysis seems to suggest that those texts provide some link between the two sections. We may infer that the botanical and recipes sections have distinct contents within a shared larger category.
        
Based on this combined qualitative and quantitative approach, we consider this strong evidence that the Voynich Manuscript is some form of enciphered human language, rather than ``meaningless'' generated text. It is the most convincing that there are distinct relationships between illustrative topics and NMF topics, while also maintaining some overlapping vocabulary between sections (see Appendix). 

Figure~\ref{assoc} provides a summary of the tf-idf topics (both the 6 topic analysis and the 40 words per page analysis), along with language, scribe, and illustrated section (per Figure~\ref{fig:vmap} above).


\begin{figure*}[ht]
    \centering
    \includegraphics[width=\linewidth]{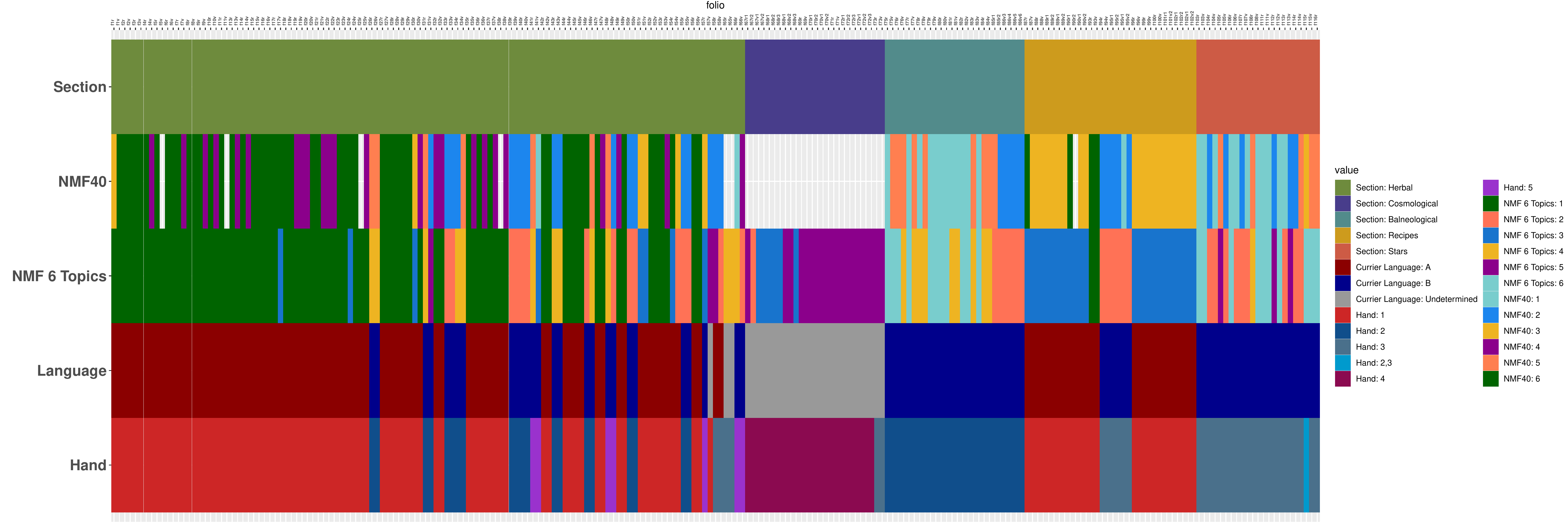}
    \caption{Summary of topics, hands, languages, and scribes}
    \label{assoc}
\end{figure*}

\bibliographystyle{plainnat} 
\bibliography{acl2015}

\appendix

\section{LDA}\label{AppA}

 \begin{figure}[ht]
    \centering
    \includegraphics[width=\linewidth]{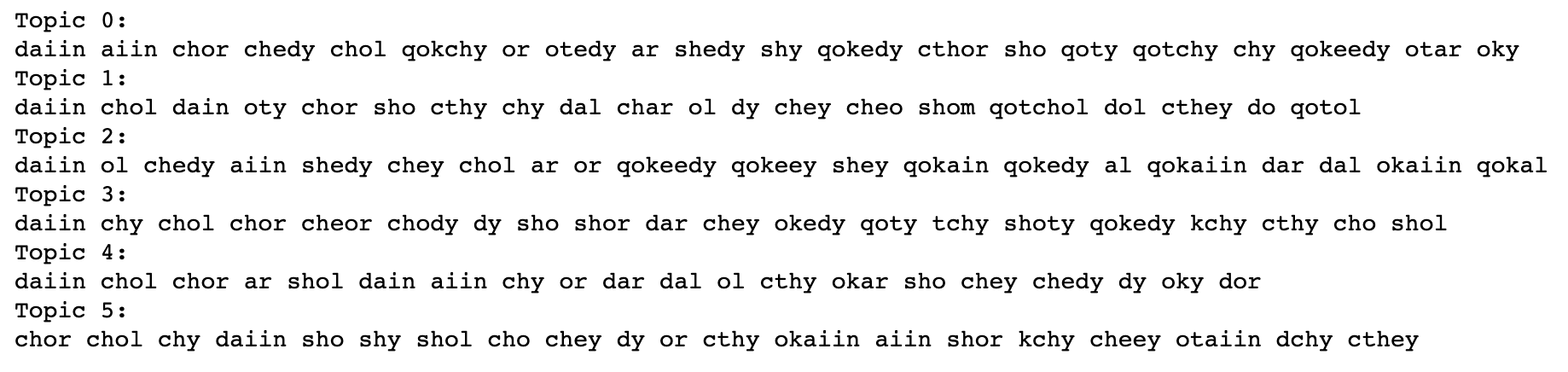}
    \caption{Top 20 words per topic for LDA, 6 topics}
\end{figure}

\section{LSA}\label{AppB}

 \begin{figure}[ht]
    \centering
    \includegraphics[width=\linewidth]{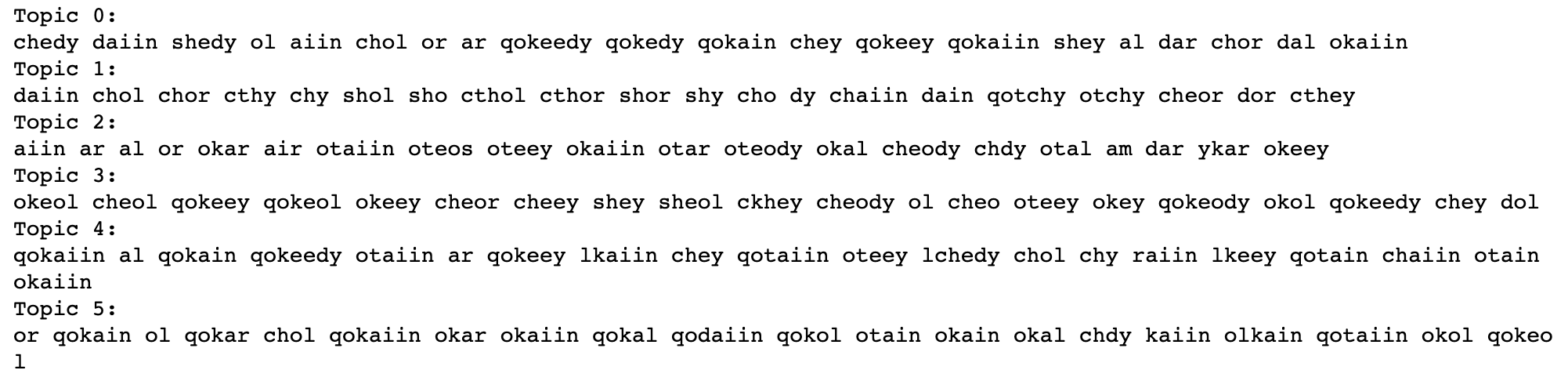}
    \caption{Top 20 words per topic for LSA, 6 topics}
\end{figure}

\section{NMF}\label{AppC}

 \begin{figure}[ht]
    \centering
    \includegraphics[width=\linewidth]{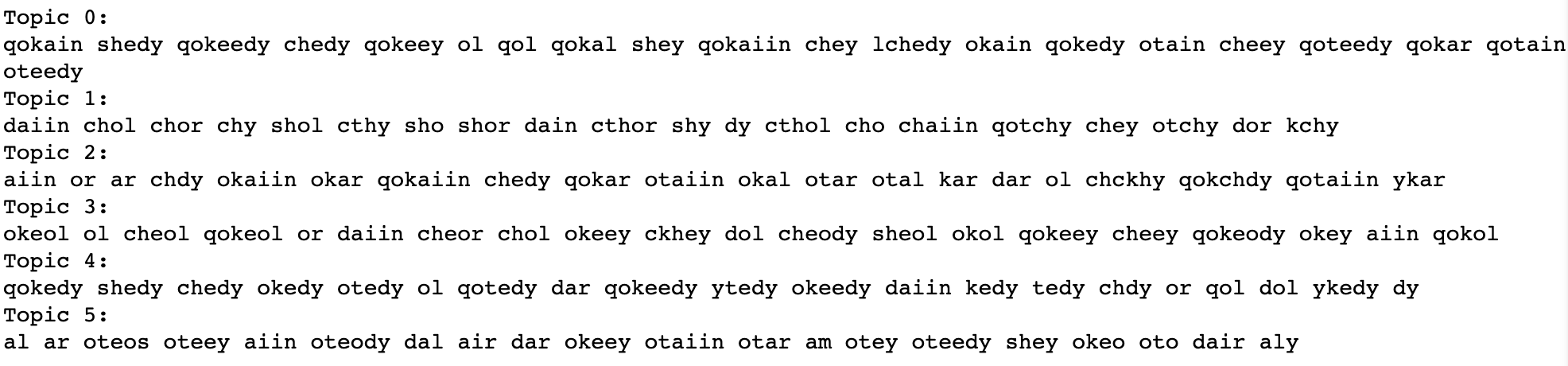}
    \caption{Top 20 words per topic for NMF, 6 topics}
\end{figure} 

 \begin{figure}[ht]
    \centering
    \includegraphics[width=\linewidth]{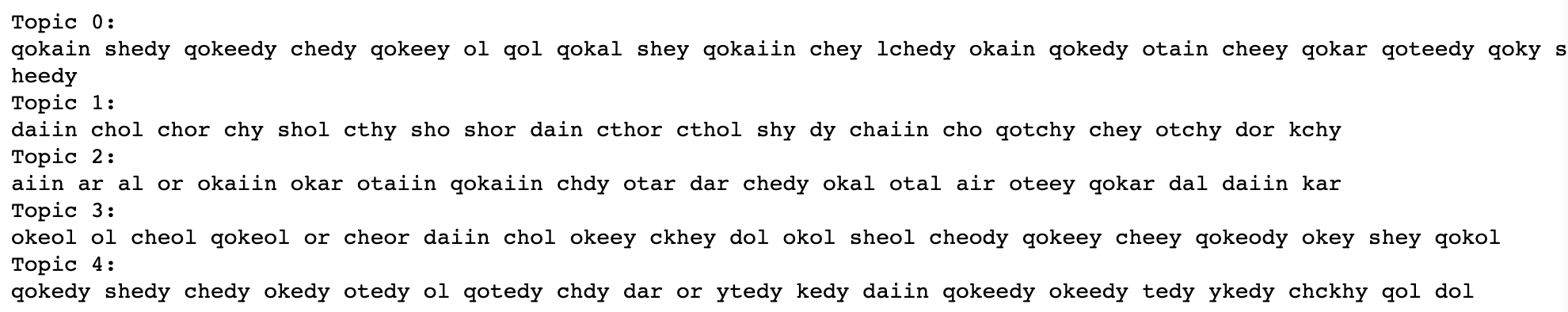}
    \caption{Top 20 words per topic for NMF, 5 topics}
\end{figure}

\end{document}